\documentclass[12pt]{article}

\usepackage[margin=1in]{geometry}
\usepackage{times}
\usepackage{xspace}
\usepackage[dvipsnames]{xcolor}
\usepackage{graphicx}
\usepackage{amsmath,amssymb}
\usepackage{booktabs}
\usepackage[numbers,sort&compress]{natbib}
\usepackage[format=plain,labelformat=simple]{caption}
\usepackage{subcaption}
\usepackage[hyphens]{url}
\usepackage{authblk}

\usepackage{amsfonts}
\usepackage{longtable}
\usepackage{multirow}
\usepackage{colortbl}
\usepackage{algorithm}
\usepackage{algorithmic}
\usepackage{enumitem}
\usepackage{bm}
\usepackage{tcolorbox}
\newcommand{\ours}{MIRAGE}
\newcommand{\method}{MIRAGE}
\newcommand{\feat}[1]{\bm{f}_{#1}}

\providecommand{\eg}{\textit{e.g.}\xspace}

\usepackage[breaklinks,colorlinks,allcolors=blue]{hyperref}
\usepackage{cleveref}

\title{MIRAGE:\@ Model-agnostic Industrial Realistic Anomaly Generation and Evaluation for Visual Anomaly Detection}

\author[1]{Jinwei Hu}
\author[2]{Francesco Borsatti}
\author[3]{Arianna Stropeni}
\author[4]{Davide Dalle Pezze}
\author[5]{Manuel Barusco}
\author[6]{Gian Antonio Susto}

\affil[ ]{\centering\normalsize University of Padova, Italy}

\affil[1]{\texttt{jinwei.hu@phd.unipd.it}}
\affil[2]{\texttt{francesco.borsatti.1@phd.unipd.it}}
\affil[3]{\texttt{arianna.stropeni@studenti.unipd.it}}
\affil[4]{\texttt{davide.dallepezze@unipd.it}}
\affil[5]{\texttt{manuel.barusco@phd.unipd.it}}
\affil[6]{\texttt{gianantonio.susto@unipd.it}}

\date{}

\begin{document}
\maketitle
\begin{abstract}
Industrial visual anomaly detection (VAD) methods are typically trained on normal samples only, yet performance improves substantially when even limited anomalous data is available.
Existing anomaly generation approaches either require real anomalous examples, demand expensive hardware, or produce synthetic defects that lack realism.
We present \ours{} (Model-agnostic Industrial Realistic Anomaly Generation and Evaluation), a fully automated pipeline for realistic anomalous image generation and pixel-level mask creation that requires no training and no anomalous images.
Our pipeline accesses any generative model as a black box via API calls, uses a VLM for automatic defect prompt generation, and includes a CLIP-based quality filter to retain only well-aligned generated images.
For mask generation at scale, we introduce a lightweight, training-free \emph{dual-branch semantic change detection} module combining text-conditioned Grounding DINO features with fine-grained YOLOv26-Seg structural features.
We benchmark four generation methods using Gemini 2.5 Flash Image (Nano Banana) as the generative backbone, evaluating performance on MVTec AD and VisA across two distinct tasks: (i) downstream anomaly segmentation and (ii) visual quality of the generated images, assessed via standard metrics (IS, IC-LPIPS) and a human perceptual study involving 31 participants and 1,550 pairwise votes. The results demonstrate that \ours{} offers a scalable, accessible foundation for anomaly-aware industrial inspection that requires no real defect data. As a final contribution, we publicly release a large-scale dataset comprising 500 image-mask pairs per category for every MVTec AD and VisA class, over 13,000 pairs in total, alongside all generation prompts and pipeline code.
\end{abstract}

\section{Introduction}
\label{sec:intro}

Visual Anomaly Detection (VAD) in industrial manufacturing aims to identify and localize defects in products using visual inspection~\cite{bergmann2019mvtec,zou2022spot}.
Most state-of-the-art approaches operate under the one-class paradigm: models learn a representation of normality from defect-free training images and flag deviations at test time~\cite{roth2022towards,defard2021padim,liu2023simplenet}.
Although effective, this paradigm can be improved with even a small number of anomalous examples~\cite{zavrtanik2021draem,zhang2024realnet} alongside the normal ones.

This observation has motivated a growing line of work on \emph{synthetic anomaly generation}: creating realistic defective images together with accurate pixel-level masks from normal data alone, so that downstream detectors can be trained in a supervised fashion without requiring real defect samples.
The underlying principle is straightforward: better synthetic anomalies (and their associated masks) translate directly into better downstream anomaly detection and segmentation performance.

However, existing generation methods face significant practical limitations.
Copy-paste approaches such as DRAEM~\cite{zavrtanik2021draem} and GLASS~\cite{chen2024unified} produce anomalies through texture blending and Perlin noise masks that often look visually unrealistic.
Diffusion-based methods like RealNet~\cite{zhang2024realnet} require training a per-category DDPM, adding computational overhead and limiting scalability.
Few-shot generation methods such as AnomalyDiffusion~\cite{hu2024anomalydiffusion}, DualAnoDiff~\cite{jin2025dualanodiff}, and Anodapter~\cite{11000123} achieve high fidelity but require real anomalous examples at training time, which are precisely the data that industrial settings lack.
The current state of the art among training-free, zero-shot methods is AnomalyAny~\cite{sun2025anomalyany}, which leverages a frozen Stable Diffusion model with attention-guided optimization.
While AnomalyAny avoids training on anomalous data, it still requires a GPU with at least 30\,GB of VRAM, long generation times, and substantial implementation effort, making it difficult to adopt in practical industrial deployments and hard to upgrade when newer generative models become available.

In this work, we take a fundamentally different approach.
Rather than engineering anomaly synthesis heuristics or adapting diffusion pipelines, we build a \emph{model-agnostic pipeline} that treats the generative model as a black box, accessed through a simple API call.
In our current instantiation, we use Gemini~2.5 Flash Image~\cite{gemini2025} (internally known as Nano Banana) as the generation backbone, but the pipeline is designed so that any current or future generative model can be swapped in with minimal effort.
Similarly, defect prompts are generated automatically by a vision-language model (VLM), which can itself be replaced as better VLMs emerge.
The entire pipeline is fully automated and unsupervised with respect to the anomalous domain: given only normal reference images, it produces photorealistic anomalous images at scale through API calls, requiring no training, no anomalous examples, no expensive hardware, and no manual prompt engineering.

\begin{figure}[!thb]
    \centering
    \includegraphics[width=0.6\linewidth]{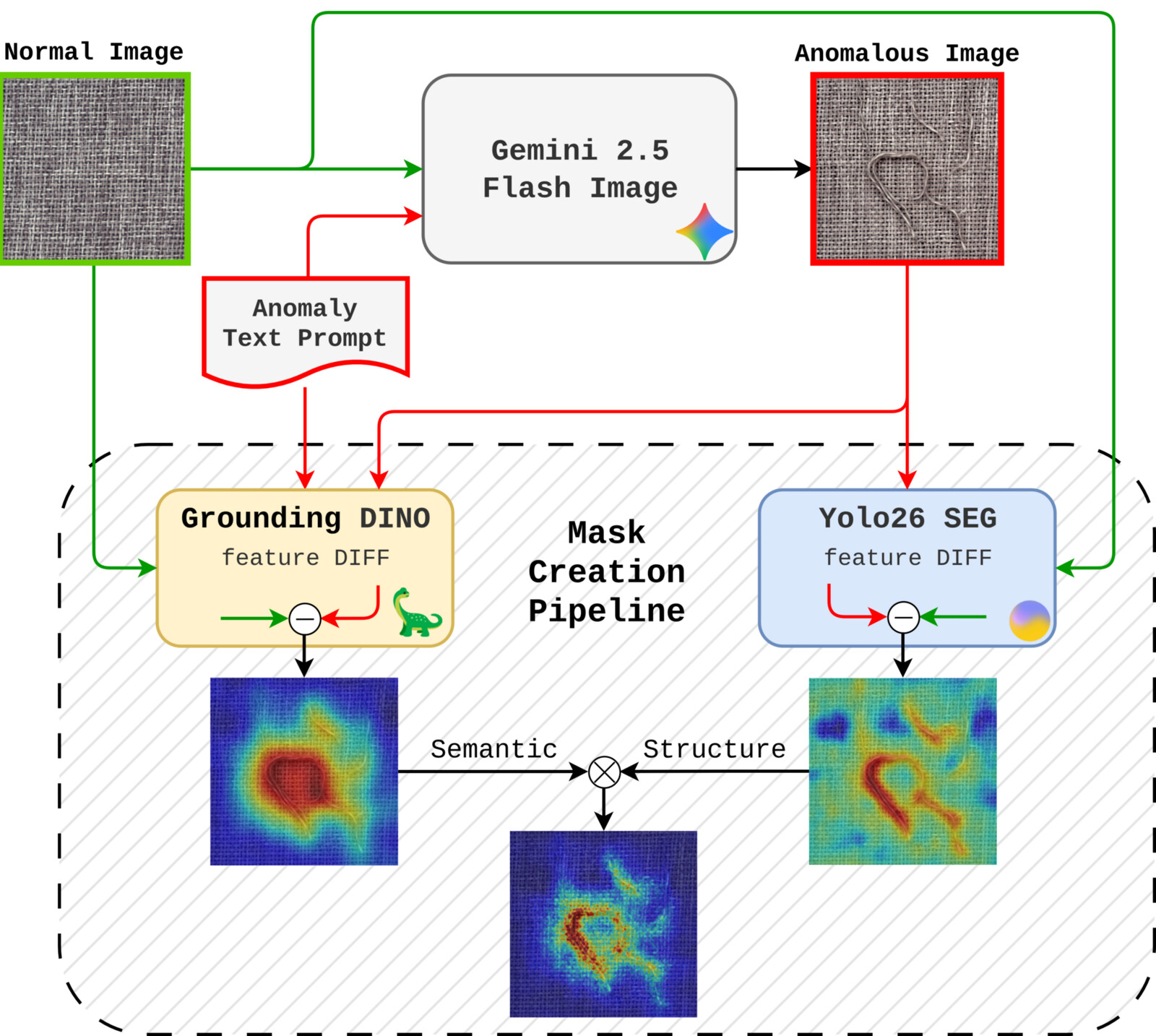}
    \caption{\textbf{Overview of \ours{}.} \textbf{Top:} A normal image and an anomaly text prompt are fed to Gemini~2.5 Flash Image, which generates a photorealistic anomalous image. \textbf{Bottom (Mask Creation Pipeline):} The normal image, anomalous image, and text prompt are processed through two branches. The \emph{semantic branch} (left) computes text-conditioned feature differences using Grounding DINO, yielding a coarse semantic anomaly map. The \emph{structural branch} (right) computes unconditional feature differences using YOLOv26-L-Seg, yielding a fine-grained structural change map. The element-wise product of both maps produces the final anomaly mask.}
    \label{fig:pipeline}
\end{figure}

A key challenge in anomaly generation is obtaining accurate pixel-level masks that delineate the generated defects.
We address this with a novel \emph{training-free semantic change detection pipeline} that, given the normal reference image, the generated anomalous image, and the defect text prompt, automatically produces a fine-grained segmentation mask (see Figure \ref{fig:pipeline}).
Our pipeline operates through dual-branch feature differencing: a \emph{semantic branch} using Grounding DINO~\cite{liu2023grounding} conditioned on the anomaly text prompt, and a \emph{structural branch} using YOLOv26-L-Seg~\cite{sapkota2025yolo26} for unconditional fine-grained change detection.
Multiplying the two heatmaps yields masks that are both spatially precise and semantically focused.

Beyond the pipeline itself, we provide what is, to our knowledge, the most extensive benchmark of anomaly generation methods for industrial VAD to date.
We conduct a comprehensive evaluation comprising: (i)~downstream anomaly segmentation using a U-Net trained on synthetic data, (ii)~automated visual quality metrics including Inception Score (IS) and Intra-Cluster Learned Perceptual Image Patch Similarity (IC-LPIPS), and (iii)~a human perceptual study with pairwise votes ranked using TrueSkill scores with tie handling.
All evaluations are conducted on both MVTec~AD and VisA, comparing four generation methods.

\noindent Our contributions are as follows:

\noindent\textbf{(1) Model-agnostic generation pipeline.} We introduce a fully automated, plug-and-play pipeline for zero-shot industrial anomaly generation that accesses generative models via API, requires no training, no anomalous images, and no costly hardware. A CLIP-based~\cite{radford2021learning} quality filter automatically selects generated images whose anomaly semantics are well aligned with the target prompt, discarding generation failures and improving downstream task performance. Swapping in a newer generative model or VLM requires changing only the API endpoint, unlike prior methods~\cite{sun2025anomalyany,zhang2024realnet} that demand significant re-implementation effort. By making the code open-source\footnote{\url{https://github.com/vadnomalous/mirage}}, we enable reproducibility and allow future researchers to adopt the framework.

\noindent\textbf{(2) Extensive and robust benchmark.} We provide a thorough comparative evaluation of four generation methods on MVTec~AD and VisA, encompassing downstream anomaly segmentation (U-Net), automated visual quality metrics (IS, IC-LPIPS), and a human perceptual study ranked via TrueSkill. Our method achieves a 67.2\% win rate, within 0.28 TrueSkill points of real images, substantially outperforming all competing generation methods.

\noindent\textbf{(3) Training-free mask generation.} We introduce a lightweight, dual-branch semantic change detection module combining Grounding DINO and YOLOv26-L-Seg feature differencing for automatic, fine-grained anomaly mask generation, achieving 0.92 pixel-level AUROC with minimal calibration effort.

\noindent\textbf{(4) Large-scale public dataset.} 
Prior works neither release datasets nor provide their generation prompts, limiting reproducibility and requiring the generation of new images from scratch, limiting the usefulness of the research that could be used directly by research that desire to work directly on supervised segmentation tasks.
Therefore, we release 500 image--mask pairs per category across \emph{all} MVTec~AD and VisA categories (over 13{,}000 pairs)\footnote{\url{https://huggingface.co/datasets/visualanom/mirage_mvtec_visa}}, together with all generation prompts. 
This is the largest such release to date, and we believe it will be valuable not only for anomaly detection research but also for related tasks such as defect classification and industrial image segmentation.

\section{Related Work}
\label{sec:related}

\paragraph{Industrial Visual Anomaly Detection.}

Industrial VAD has seen steady progress, starting from renowned methods suchas PatchCore~\cite{roth2022towards}, PaDiM~\cite{defard2021padim}, and CFA~\cite{lee2022cfa}, which construct a memory bank to model normality and detect out-of-distribution patches.
Reconstruction-based methods include normalizing flows~\cite{gudovskiy2022cflow}, and student-teacher frameworks~\cite{bergmann2020uninformed,deng2022anomaly}, learn to reconstruct normal images and flag high reconstruction error as anomalous.
Both paradigms are inherently limited by the absence of anomalous supervision.
Recent work has shown that even a small number of labeled anomalous samples, real or synthetic, can significantly boost detection and segmentation performance~\cite{zavrtanik2021draem,zhang2024realnet}, motivating the study of anomaly generation as a means to close this gap.

\paragraph{Zero-Shot Anomaly Generation.}

A first family of anomaly generation methods operates without any real anomalous examples.
DRAEM~\cite{zavrtanik2021draem} pioneered the synthetic anomaly paradigm by pasting Perlin noise-textured patches onto normal images and training a reconstruction-discrimination network; while effective, its noise-based anomalies lack the appearance of real defects.
GLASS~\cite{chen2024unified} unifies feature-level synthesis via gradient-ascent-guided Gaussian noise with image-level Perlin-mask texture overlay, achieving strong results especially on weak defects, but its synthetic anomalies still differ visibly from real-world defect appearances.
RealNet~\cite{zhang2024realnet} introduces Strength-controllable Diffusion Anomaly Synthesis (SDAS), which trains a per-category Denoising Diffusion Probabilistic Model (DDPM) on normal images and generates anomalous images by perturbing the reverse diffusion process with a controllable strength parameter; however, training a separate diffusion model per category adds high computational cost.
AnomalyAny~\cite{sun2025anomalyany} is the state of the art for zero-shot anomaly generation, and it leverages a frozen pretrained Stable Diffusion model with attention-guided optimization for unseen anomaly generation without additional training.
While AnomalyAny represents the current state of the art among zero-shot methods, it requires a GPU with at least 30\,GB of VRAM, entails long generation times, and its tight coupling to a specific diffusion pipeline makes it difficult to upgrade when newer generative models become available, all of which limit its practical deployment in industrial settings.

Most of the above methods evaluate generation quality primarily through the Inception Score (IS), which measures distributional properties but may not capture the human-perceived realism of individual samples, the same metric used by the current state of the art\cite{sun2025anomalyany}.
In contrast, our work differs in using a general-purpose generative model accessed via API and validating quality through a comprehensive benchmark that includes human perceptual studies and IC-LPIPS alongside IS.

\paragraph{Anomaly Generation for VAD.}

A prominent paradigm uses synthetic anomalies to train a discriminator or anomaly detector, sharpening the decision boundary between normal and anomalous representations.
SimpleNet~\cite{liu2023simplenet} exemplifies this by generating anomalies directly in the latent feature space to supervise a compact discriminative head.
As studied in GLASS~\cite{chen2024unified}, the quality of synthetic anomalies directly governs the tightness of the learned normal-vs-abnormal boundary, motivating the pursuit of more realistic synthesis.
A second line of work~\cite{hu2024anomalydiffusion, jin2025dualanodiff, 11000123, xu2025stage} conditions generative models on a small set of real defect examples to produce highly in-distribution anomalies. 
While this yields high visual fidelity, it presupposes anomalous supervision and real anomalous data that is often unavailable in practice.
Our method operates in the zero-shot regime alongside GLASS, RealNet, and AnomalyAny: it requires \emph{no anomalous images at any stage}, instead uses a VLM to propose plausible defect types from normal images alone and a general-purpose generative model to synthesize anomalies purely from text. This offers significant practical advantages in simplicity, and upgradeability.

\paragraph{Change Detection via Feature Differencing.}

Change detection identifies meaningful differences between two images of the same scene.
Classical methods rely on pixel-level differencing or handcrafted features~\cite{radke2005image}, while modern approaches extract features from pretrained networks and compute differences in embedding space~\cite{chen2020spatial,fang2023changer}.
View-Delta~\cite{sachdeva2023change} performs semantic change detection using foundation model features but was designed for natural scene changes (\eg, construction, vegetation growth) and produces coarse change maps insufficient for industrial defect segmentation; it also requires training on change detection datasets.
\\
\\
Our approach draws on the feature differencing paradigm but introduces two key novelties: (i)~text-conditioned semantic differencing to suppress false positives from non-defect variations, and (ii)~a dual-branch architecture combining semantic and structural features without any training, making it immediately applicable to arbitrary industrial products.

\section{Methodology}\label{sec:method}

Our pipeline consists of three stages: (1)~anomalous image generation via prompted multimodal generation, (2)~CLIP-based quality filtering of generated images, and (3)~automatic mask creation through dual-branch semantic change detection.
A key design principle is \emph{model agnosticism}: both the generative model and the VLM are accessed as black boxes through API calls, so that upgrading to a newer or better model requires only changing the endpoint, with no architectural modifications or retraining.
An overview is shown in \cref{fig:pipeline}.

\subsection{Anomaly Prompt Generation}\label{sec:prompt_gen}

Given a set of normal reference images $\{I_n^{(k)}\}_{k=1}^{K}$ for a product category (\eg, $K{=}5$), we query a vision-language model (VLM) to propose plausible defect types.
The choice of VLM is not prescribed by our pipeline; any model with visual understanding and text generation capabilities can be used, we specifically used ChatGPT 5.
We provide the VLM with the reference images and a structured prompt requesting a list of realistic manufacturing defect types with concise descriptions suitable for conditioning an image generation model.

\noindent The VLM returns a structured list of defect descriptions $\{d_j\}_{j=1}^{10}$, each consisting of a short defect name and a descriptive sentence (\eg, \textit{``surface scratch: a thin, shallow linear mark across the surface, exposing a slightly lighter layer underneath''}).
This automated prompt generation eliminates the need for manual domain expertise while producing defect descriptions grounded in the visual appearance of the actual product.
Because the VLM operates solely on normal images, the entire prompt generation process is fully unsupervised with respect to the anomalous domain.

\subsection{Anomalous Image Generation}\label{sec:image_gen}

For each normal image $I_n$ and defect description $d_j$, we generate an anomalous image using a general-purpose generative model accessed via API.\@
In our current instantiation, we use Gemini~2.5 Flash Image~\cite{gemini2025} (Nano Banana), but the pipeline is designed to work with any generative model that accepts an image and a text prompt as input.
The generation is conditioned on both the normal image and a structured text prompt instructing the model to produce a photorealistic version with the specified defect while preserving background, lighting, and overall appearance. All prompts used for generation are provided in the released code and supplementary material.

\noindent Let $G$ denote the generative model; then, the anomalous image is:
\begin{equation}
    I_a = G(I_n, d_j).
    \label{eq:generation}
\end{equation}
By conditioning on the original normal image, the generative model preserves the overall scene layout, illumination, and texture while introducing the specified defect with high fidelity.
The entire generation process requires only API calls: no local GPU resources, no model weights to download, and no training of any kind.

For each category, we generate 50 anomalous images per defect type (10 defect types $\times$ 50 images = 500 per category), sampling normal images uniformly from the training set.

\subsection{CLIP-Based Quality Filtering}\label{sec:clip_filter}

Since the generative model occasionally fails to introduce the requested anomaly or produces artifacts unrelated to the target defect, we employ a CLIP-based quality filter to automatically select well-aligned generated images.
Given a normal image $I_n$, its corresponding anomalous image $I_a$, the normal prompt $p_n$ (\eg, \textit{``a normal [product]''}), and the anomaly prompt $p_a$ (the defect description $d_j$), we compute four CLIP image--text similarities and impose three conditions:

\noindent\textbf{(C1)}~$\text{sim}(I_a, p_a) \geq \text{sim}(I_n, p_n)$: the anomalous image should align with its anomaly prompt at least as well as the normal image aligns with the normal prompt, ensuring the generated image is semantically coherent.

\noindent\textbf{(C2)}~$\text{sim}(I_a, p_a) \geq \text{sim}(I_a, p_n)$: the anomalous image should align more closely with the anomaly prompt than with the normal prompt, confirming the presence of the intended defect.

\noindent\textbf{(C3)}~$\text{sim}(I_a, p_a) \geq \text{sim}(I_n, p_a)$: the anomalous image should align more closely with the anomaly prompt than the normal image does, verifying that the defect is present in the generated image and absent in the original.

\noindent Images that satisfy all three conditions are retained; the rest are discarded, this ensures more alignment with the generated images and the prompts.
This lightweight filter adds negligible computational overhead (a single CLIP forward pass per image) and, as shown in \cref{sec:clip_selection}, consistently improves downstream task performance by removing generation failures and poorly aligned samples.

\subsection{Dual-Branch Semantic Change Detection}\label{sec:change_detection}

The challenge is to obtain a pixel-level anomaly mask $M$ given the normal image $I_n$, the generated anomalous image $I_a$, and the defect prompt $d_j$, \emph{without any training}.
Na\"ive pixel differencing fails due to the subtle global variations introduced by the generative model (\eg, slight color shifts, texture re-rendering).
We propose a dual-branch approach that operates in learned feature space, combining semantic and structural cues.

\paragraph{Semantic Branch: Grounding DINO Feature Diff.}\label{sec:semantic}

Grounding DINO~\cite{liu2023grounding} is an open-set object detector that fuses visual and textual features through cross-modal attention.
We use the Grounding DINO Tiny variant\footnote{\texttt{IDEA-Research/grounding-dino-tiny}} and exploit its intermediate feature representations, which encode both spatial and semantic information conditioned on a text query.

Let $\phi_{\text{GD}}(I, t)$ denote the feature map extracted from a selected intermediate layer of Grounding DINO when processing image $I$ with text prompt $t$.
We extract a few characterizing keywords $t_j$ from the defect description $d_j$ (\eg, for the description \textit{``surface scratch: a thin, shallow linear mark\ldots''}, we use $t_j = $ \textit{``scratch, linear mark''}).

We compute feature maps for both images conditioned on the \emph{same anomaly prompt}:
\begin{align}
    \feat{n}^{\text{sem}} &= \phi_{\text{GD}}(I_n, t_j), \label{eq:feat_normal_sem} \\
    \feat{a}^{\text{sem}} &= \phi_{\text{GD}}(I_a, t_j). \label{eq:feat_anom_sem}
\end{align}

The rationale for conditioning the normal image with the anomaly prompt is to capture the ``base activation'': the response of the text-conditioned features to the normal image when queried about the defect.
Subtracting this baseline removes false positives arising from textures or structures in the normal image that partially match the defect description:
\begin{equation}
    S_{\text{sem}} = \left\| \feat{a}^{\text{sem}} - \feat{n}^{\text{sem}} \right\|_2,
    \label{eq:sem_diff}
\end{equation}
where $S_{\text{sem}} \in \mathbb{R}^{H' \times W'}$ is the semantic anomaly score map and $\|\cdot\|_2$ denotes the $L_2$ norm computed across the channel dimension.
This map is upsampled to the original image resolution via bilinear interpolation:
\begin{equation}
    \hat{S}_{\text{sem}} = \text{Upsample}(S_{\text{sem}}, H, W).
    \label{eq:sem_upsample}
\end{equation}

The semantic branch produces a coarse anomaly localization that is highly responsive to the described defect type but lacks the spatial precision needed for pixel-accurate segmentation.

\paragraph{Structural Branch: YOLOv26-L-Seg Feature Diff.}\label{sec:structural}

To capture fine-grained structural changes, we leverage the multi-scale feature pyramid of YOLOv26-L-Seg\footnote{\texttt{openvision/yolo26-l-seg}}~\cite{sapkota2025yolo26}, a real-time segmentation model whose backbone produces rich spatially-detailed features well-suited for pixel-level change detection.
We choose the segmentation variant over the detection-only model because its feature maps retain finer spatial resolution, yielding higher-quality anomaly masks.
The structural branch operates \emph{without text conditioning}, directly comparing visual features between the normal and anomalous images.

Let $\psi_l(I)$ denote the feature map from the $l$-th layer of the YOLOv26-L-Seg backbone for image $I$.
We compute per-layer differences across $L$ selected layers:
\begin{equation}
    S_{\text{str}}^{(l)} = \left\| \psi_l(I_a) - \psi_l(I_n) \right\|_2, \quad l = 1, \ldots, L.
    \label{eq:str_diff}
\end{equation}

Each per-layer score map is upsampled to the original resolution and normalized to $[0, 1]$:
\begin{equation}
    \hat{S}_{\text{str}}^{(l)} = \text{Normalize}\!\left(\text{Upsample}(S_{\text{str}}^{(l)}, H, W)\right).
    \label{eq:str_upsample}
\end{equation}

The multi-scale structural score map is obtained by averaging across layers:
\begin{equation}
    \hat{S}_{\text{str}} = \frac{1}{L} \sum_{l=1}^{L} \hat{S}_{\text{str}}^{(l)}.
    \label{eq:str_avg}
\end{equation}

This multi-scale aggregation captures both fine local changes (from early layers) and broader structural deformations (from deeper layers), providing spatially precise change detection regardless of defect semantics.

\paragraph{Mask Fusion and Calibration.}\label{sec:fusion}

The final anomaly score map is the element-wise product of the normalized semantic and structural maps:
\begin{equation}
    S = \hat{S}_{\text{sem}} \odot \hat{S}_{\text{str}},
    \label{eq:fusion}
\end{equation}
where $\odot$ denotes the Hadamard (element-wise) product.
This multiplicative fusion ensures that the final mask highlights regions that exhibit \emph{both} semantic relevance (consistent with the described defect type) \emph{and} structural change (actual pixel-level modification), effectively suppressing false positives from either branch in isolation.

To convert the continuous score map $S$ into a binary mask, we determine an optimal threshold $\tau^*$ per category using a calibration set.
We generate 5--8 high-quality reference masks per defect category using Gemini~2.5 Flash Image V3, which produces accurate masks but at substantially higher cost.
The threshold is selected to maximize the pixel-level AUROC (Area Under the Receiver Operating Characteristic curve) on this calibration set:
\begin{equation}
    \tau^* = \arg\max_{\tau} \; \text{AUROC}\!\left(\mathbf{1}[S > \tau], \; M_{\text{ref}}\right),
    \label{eq:threshold}
\end{equation}
where $M_{\text{ref}}$ denotes the V3-generated reference masks.
The final binary mask is:
$M = \mathbf{1}[S > \tau^*].$
Once calibrated, the mask creation pipeline processes each image in a single forward pass through both branches, requiring no iterative refinement or expensive generative model calls.
The entire dual-branch pipeline runs in approximately 1~second per image (non-batched) using roughly 3\,GB of VRAM on a single NVIDIA RTX A6000, making it practical for large-scale dataset generation.

\section{Experimental Setup}
\label{sec:experiments}

A central contribution of this work is a robust benchmark of anomaly generation methods for industrial VAD.
Prior works typically evaluate generation quality through metrics such as IS (Inception Score), we also conduct an evaluation comprising downstream task performance, and human perceptual assessment, all performed on two standard benchmarks across four generation methods.

\paragraph{Benchmarks.}
We evaluate on two standard industrial VAD benchmarks: MVTec AD~\cite{bergmann2019mvtec} (15 categories, 5 texture and 10 object classes) and VisA~\cite{zou2022spot} (12 categories of complex industrial objects).
The test sets of both datasets include real normal and real anomalous images, and provide image-level classification labels as well as pixel-level anomaly annotations, enabling a comprehensive evaluation of both detection and segmentation performance.

\paragraph{Compared methods.}
We compare against three recent anomaly generation approaches that, like ours, do not require real anomalous examples:
(1)~\textbf{GLASS}~\cite{chen2024unified},
(2)~\textbf{RealNet} (SDAS)~\cite{zhang2024realnet}, we adopt SDAS as it is the most competitive generation baseline;
(3)~\textbf{AnomalyAny}~\cite{sun2025anomalyany}, the current state of the art among zero-shot anomaly generation.
For each method, we use the official code and default hyperparameters to generate anomalous training data.

\paragraph{Downstream VAD and segmentation.}
To evaluate the utility of generated data for downstream anomaly segmentation, we train a binary segmentation U-Net~\cite{ronneberger2015unet} on the synthetic anomalous images and their corresponding generated masks, together with defect-free images from the original datasets labeled as normal. From all methods, we train on 100 image-mask pairs per category, balanced with normal images. 
The model is trained for 100 epochs with a learning rate of $10^{-4}$ and evaluated on the official test splits of the respective dataset MVTec~AD, and VisA.
This setup directly tests the practical value of each generation method: better synthetic data should translate into better real-world segmentation performance.

\paragraph{Evaluation metrics.}
We report: (i)~Image-level AUROC for anomaly detection, (ii)~Pixel-level AUROC for anomaly segmentation (i.e.\ computed on the flattened anomaly map, treating each pixel as an independent prediction), (iii)~Inception Score (IS; higher is better), which reflects image quality via both semantic correctness and visual sharpness, (iv)~Intra-Cluster LPIPS (IC-LPIPS), which measures perceptual alignment to real defects within each defect cluster (higher indicates closer perceptual match), and (v)~Human perceptual realism from a controlled pairwise comparison study ranked using TrueSkill scores~\cite{graepel2007bayesian} (computed via the relative Python package with tie handling) and win rates.

\section{Results}

\subsection{Generation Quality Evaluation}
\label{sec:exp_quality}
\begin{figure*}
    \centering
    \includegraphics[width=0.9\linewidth]{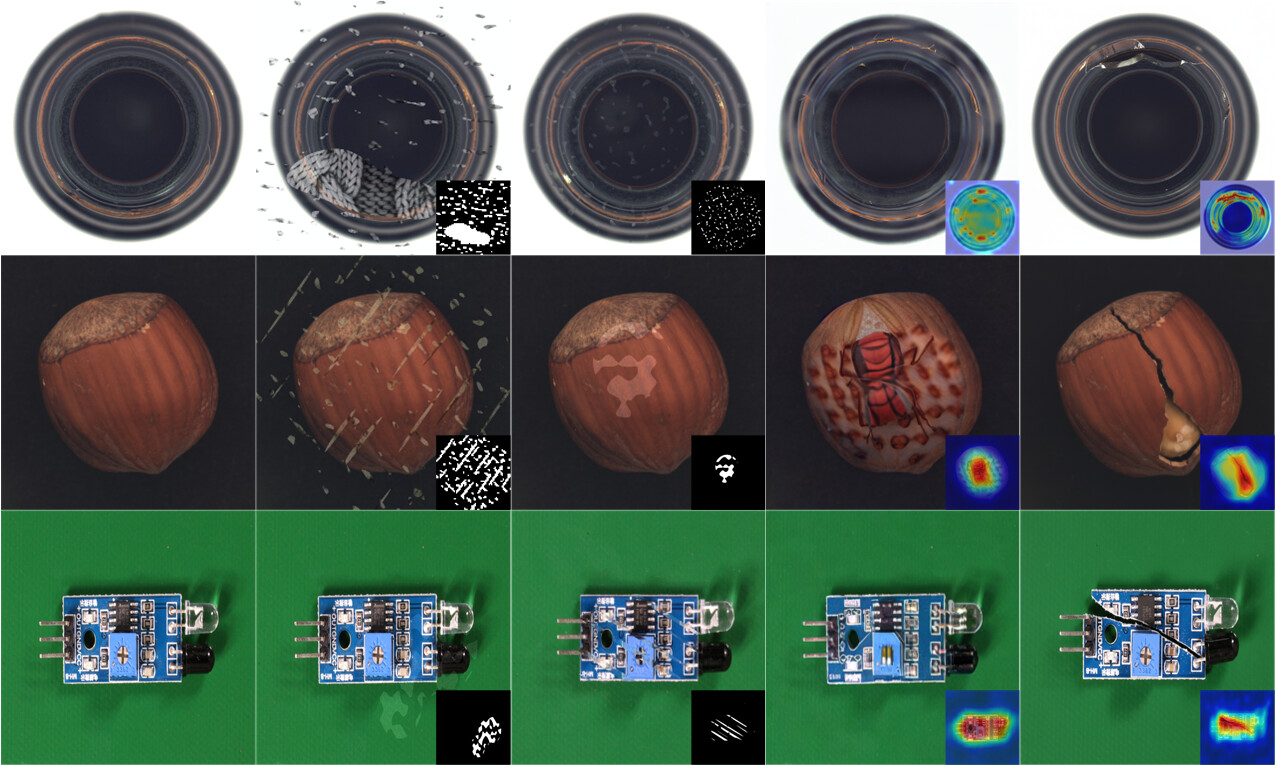}
    \caption{\textbf{Examples of synthetic anomalies generated.} From left to right: (1) original input image, (2) GLASS, (3) RealNet, (4) AnomalyAny, and (5) MIRAGE. The rows represent three object classes: bottle (top row), hazelnuts (mid row) and electronic sensor modules (bottom row).
    The relative generated anomaly map or mask is overlaid on the bottom right of each image.}
    \label{fig:generated_images}
\end{figure*}

\paragraph{Human Perceptual Study}

We conduct a human evaluation study with 31 participants to assess the visual realism of generated anomalous images across methods.
In each trial, an evaluator is shown two anomalous images from different methods (or real) and asked to select the more realistic one, or to indicate a tie.
Images are randomly sampled across categories and methods, with all comparisons blinded and randomized.
We aggregate 1{,}550 collected votes into a global ranking using the TrueSkill rating system~\cite{graepel2007bayesian}, reporting TrueSkill scores ($\mu \pm \sigma$) alongside win rates (excluding ties from the denominator).

\begin{table}[t]
    \centering
    \caption{\textbf{Human perceptual realism evaluation.} TrueSkill ranking~\cite{graepel2007bayesian} ($\mu \pm \sigma$, higher is better), win rate (\%, excluding ties), and number of appearances across all pairwise comparisons (31 participants, 1{,}550 total votes). Real images are included as an upper bound.}
    \label{tab:human_eval}
    \small
    \begin{tabular}{lccc}
        \toprule
        Method & TrueSkill ($\mu \pm \sigma$) $\uparrow$ & Win\% $\uparrow$ & App. \\
        \midrule
        Real images            & $28.61 \pm 0.80$ & 73.8 & 644 \\ \hline
        \textbf{\method{} (Ours)} & $28.33 \pm 0.80$ & 67.2 & 622 \\
        AnomalyAny~\cite{sun2025anomalyany} & $27.16 \pm 0.79$ & 59.2 & 613 \\
        RealNet~\cite{zhang2024realnet} & $23.54 \pm 0.79$ & 33.7 & 642 \\
        GLASS~\cite{chen2024unified} & $20.26 \pm 0.84$ & 14.4 & 605 \\
        \bottomrule
    \end{tabular}
\end{table}

\Cref{tab:human_eval} reports the results of the human perceptual study.
\method{} achieves a TrueSkill score of $28.33 \pm 0.80$, the closest to real images ($28.61 \pm 0.80$) among all generation methods, with a gap of only $0.28$ points.
In terms of win rate, \method{} wins 67.2\% of its decisive comparisons, substantially outperforming AnomalyAny (59.2\%), RealNet (33.7\%), and GLASS (14.4\%).
These results confirm that our pipeline produces significantly more realistic anomalous images than all competing methods, with a perceptual quality approaching that of real defect images (see Figure \ref{fig:generated_images} for a visual comparison among the methods for some categories). 

\begin{table}[t]
    \centering
    \caption{\textbf{Visual quality metrics on MVTec AD.} Inception Score (IS, higher is better) and Intra-Cluster LPIPS (IC-LPIPS, denoted IC-L) for generated anomalous images from each method, compared against real anomalous test images.}%
    \label{tab:visual_metrics}
    \small
    \resizebox{0.7\linewidth}{!}{%
    \begin{tabular}{l c >{\columncolor{lightgray!20}}c c >{\columncolor{lightgray!20}}c c >{\columncolor{lightgray!20}}c c >{\columncolor{lightgray!20}}c}
        \toprule
        & \multicolumn{2}{c}{AnomalyAny} & \multicolumn{2}{c}{RealNet} & \multicolumn{2}{c}{GLASS} & \multicolumn{2}{c}{\textbf{\method{}}} \\
        \cmidrule(lr){2-3} \cmidrule(lr){4-5} \cmidrule(lr){6-7} \cmidrule(lr){8-9}
        Category & IS & IC-L & IS & IC-L & IS & IC-L & IS & IC-L \\
        \midrule
        Bottle        & 1.73 & 0.17 & 1.58 & 0.16 & 2.08 & 0.26 & 2.31 & 0.23 \\
        Cable         & 2.06 & 0.41 & 1.65 & 0.18 & 1.81 & 0.42 & 2.63 & 0.53 \\
        Capsule       & 2.16 & 0.23 & 1.62 & 0.41 & 2.50 & 0.33 & 2.27 & 0.20 \\
        Carpet        & 1.10 & 0.34 & 1.03 & 0.25 & 1.29 & 0.27 & 2.09 & 0.33 \\
        Grid          & 2.31 & 0.38 & 2.24 & 0.37 & 2.46 & 0.41 & 2.80 & 0.56 \\
        Hazelnut      & 2.55 & 0.32 & 2.21 & 0.31 & 2.70 & 0.44 & 3.27 & 0.30 \\
        Leather       & 2.26 & 0.41 & 1.67 & 0.36 & 2.76 & 0.28 & 4.11 & 0.48 \\
        Metal Nut     & 1.82 & 0.27 & 1.67 & 0.24 & 1.93 & 0.44 & 2.82 & 0.42 \\
        Pill          & 2.91 & 0.30 & 1.45 & 0.26 & 2.08 & 0.38 & 2.01 & 0.28 \\
        Screw         & 1.33 & 0.32 & 1.20 & 0.31 & 1.37 & 0.48 & 1.21 & 0.30 \\
        Tile          & 2.66 & 0.53 & 1.57 & 0.48 & 1.84 & 0.36 & 2.66 & 0.50 \\
        Toothbrush    & 1.64 & 0.22 & 1.19 & 0.18 & 1.77 & 0.42 & 2.09 & 0.23 \\
        Transistor    & 1.66 & 0.28 & 1.49 & 0.30 & 1.70 & 0.35 & 1.67 & 0.40 \\
        Wood          & 1.93 & 0.41 & 2.22 & 0.41 & 2.72 & 0.44 & 4.95 & 0.51 \\
        Zipper        & 2.14 & 0.33 & 1.88 & 0.24 & 2.59 & 0.31 & 3.37 & 0.36 \\
        \midrule
        \textbf{Mean} & 2.10 & 0.37 & 1.64 & 0.30 & 2.02 & 0.33 & \textbf{2.68} & \textbf{0.38} \\
        \bottomrule
    \end{tabular}
    }
\end{table}

\paragraph{Inception Score and IC-LPIPS}

\Cref{tab:visual_metrics} reports the Inception Score (IS) and Intra-Cluster LPIPS (IC-LPIPS) for generated images from all four methods.
\method{} achieves the highest mean IS (2.68), indicating consistently higher-quality generations across categories.
Moreover, \method{} attains the best mean IC-LPIPS (0.38), suggesting that our generated defects are not only diverse but also perceptually aligned with real defect appearances within the same cluster.
Together, these metrics corroborate the human study and confirm that \method{} produces realistic, category-consistent anomalies rather than spurious artifacts.

\paragraph{Mask Quality Evaluation}
\label{sec:exp_mask}

We assess the accuracy of our change-detection pipeline by comparing its output heatmaps against pseudo-ground-truth masks obtained from Gemini~2.5 Flash Image V3, using pixel-level AUROC as the evaluation criterion.
Reference masks are collected for 1{,}000 image pairs on MVTec~AD and 720 on VisA.
Our pipeline substantially outperforms View-Delta~\cite{sachdeva2023change} on both benchmarks, achieving an average pixel-level AUROC of 0.93 compared to 0.73.

\subsection{Downstream Anomaly Segmentation}
\label{sec:exp_downstream}

The primary evaluation measures whether synthetic data improves real-world anomaly segmentation, directly testing the practical value proposition of each generation method.
We train a U-Net on the generated image--mask pairs from each method (plus normal images) and evaluate on the real test sets.
\Cref{tab:downstream} reports results on MVTec AD and VisA respectively.

\begin{table*}[t]
    \centering
    \caption{\textbf{Downstream anomaly segmentation.} Image-level AUROC (I) and pixel-level AUROC (P) for a U-Net trained on synthetic data from each generation method. Best results in \textbf{bold}. (a)~MVTec~AD. (b)~VisA.}
    \label{tab:downstream}
    \small
    \begin{minipage}[t]{0.49\textwidth}
        \resizebox{\textwidth}{!}{%
        \begin{tabular}{l c >{\columncolor{lightgray!20}}c c >{\columncolor{lightgray!20}}c c >{\columncolor{lightgray!20}}c c >{\columncolor{lightgray!20}}c}
            \toprule
            \textbf{(a)} & \multicolumn{2}{c}{AnomAny} & \multicolumn{2}{c}{RealNet} & \multicolumn{2}{c}{GLASS} & \multicolumn{2}{c}{\textbf{Ours}} \\
            \cmidrule(lr){2-3} \cmidrule(lr){4-5} \cmidrule(lr){6-7} \cmidrule(lr){8-9}
            \textbf{MVTec AD} & I & P & I & P & I & P & I & P \\
            \midrule
            Bottle      & 0.78 & 0.85 & 0.92 & 0.91 & 0.91 & 0.83 & 0.93 & 0.92 \\
            Cable       & 0.49 & 0.76 & 0.41 & 0.68 & 0.74 & 0.80 & 0.71 & 0.86 \\
            Capsule     & 0.57 & 0.89 & 0.49 & 0.88 & 0.43 & 0.87 & 0.45 & 0.93 \\
            Carpet      & 0.39 & 0.90 & 0.45 & 0.86 & 0.56 & 0.98 & 0.82 & 0.99 \\
            Grid        & 0.57 & 0.74 & 0.82 & 0.63 & 0.93 & 0.98 & 0.54 & 0.90 \\
            Hazelnut    & 0.81 & 0.96 & 0.83 & 0.90 & 0.73 & 0.92 & 0.92 & 0.99 \\
            Leather     & 0.86 & 0.97 & 0.98 & 0.96 & 0.99 & 0.99 & 0.93 & 0.96 \\
            Metal Nut   & 0.71 & 0.75 & 0.58 & 0.75 & 0.74 & 0.73 & 0.96 & 0.96 \\
            Pill        & 0.85 & 0.87 & 0.94 & 0.97 & 0.85 & 0.93 & 0.85 & 0.89 \\
            Screw       & 0.10 & 0.89 & 0.95 & 0.91 & 0.10 & 0.94 & 0.96 & 0.95 \\
            Tile        & 0.87 & 0.94 & 0.98 & 0.95 & 1.00 & 0.99 & 0.93 & 0.98 \\
            Toothbrush  & 0.65 & 0.85 & 0.58 & 0.85 & 0.79 & 0.87 & 0.87 & 0.96 \\
            Transistor  & 0.48 & 0.63 & 0.85 & 0.64 & 0.87 & 0.67 & 0.77 & 0.75 \\
            Wood        & 0.95 & 0.86 & 0.98 & 0.88 & 0.93 & 0.91 & 0.95 & 0.86 \\
            Zipper      & 0.71 & 0.84 & 0.99 & 0.86 & 0.88 & 0.95 & 0.47 & 0.89 \\
            \midrule
            \textbf{Mean} & 0.65 & 0.85 & 0.78 & 0.84 & 0.76 & 0.89 & \textbf{0.81} & \textbf{0.92} \\
            \bottomrule
        \end{tabular}%
        }
        \label{tab:downstream_mvtec}
    \end{minipage}\hfill
    \begin{minipage}[t]{0.49\textwidth}
        \resizebox{\textwidth}{!}{%
        \begin{tabular}{l c >{\columncolor{lightgray!20}}c c >{\columncolor{lightgray!20}}c c >{\columncolor{lightgray!20}}c c >{\columncolor{lightgray!20}}c}
            \toprule
            \textbf{(b)} & \multicolumn{2}{c}{AnomAny} & \multicolumn{2}{c}{RealNet} & \multicolumn{2}{c}{GLASS} & \multicolumn{2}{c}{\textbf{Ours}} \\
            \cmidrule(lr){2-3} \cmidrule(lr){4-5} \cmidrule(lr){6-7} \cmidrule(lr){8-9}
            \textbf{VisA} & I & P & I & P & I & P & I & P \\
            \midrule
            Candle      & 0.56 & 0.71 & 0.76 & 0.62 & 0.84 & 0.90 & 0.82 & 0.92 \\
            Capsules    & 0.49 & 0.73 & 0.80 & 0.87 & 0.83 & 0.93 & 0.83 & 0.89 \\
            Cashew      & 0.19 & 0.91 & 0.97 & 0.90 & 0.25 & 0.89 & 0.97 & 0.97 \\
            Chewinggum  & 0.91 & 0.99 & 0.90 & 0.92 & 0.94 & 1.00 & 0.72 & 0.97 \\
            Fryum       & 0.81 & 0.91 & 0.78 & 0.87 & 0.50 & 0.93 & 0.81 & 0.95 \\
            Macaroni1   & 0.76 & 0.94 & 0.33 & 0.90 & 0.94 & 0.99 & 0.68 & 0.95 \\
            Macaroni2   & 0.58 & 0.91 & 0.54 & 0.89 & 0.78 & 0.99 & 0.65 & 0.93 \\
            PCB1        & 0.54 & 0.69 & 0.41 & 0.75 & 0.39 & 0.74 & 0.36 & 0.77 \\
            PCB2        & 0.78 & 0.85 & 0.71 & 0.80 & 0.73 & 0.86 & 0.80 & 0.92 \\
            PCB3        & 0.31 & 0.85 & 0.58 & 0.85 & 0.42 & 0.86 & 0.63 & 0.89 \\
            PCB4        & 0.58 & 0.85 & 0.71 & 0.80 & 0.80 & 0.93 & 0.85 & 0.91 \\
            Pipe Fryum  & 0.52 & 0.94 & 0.63 & 0.92 & 0.68 & 0.89 & 0.71 & 0.99 \\
            \midrule
            \textbf{Mean} & 0.59 & 0.86 & 0.68 & 0.84 & 0.67 & 0.91 & \textbf{0.74} & \textbf{0.92} \\
            \bottomrule
        \end{tabular}%
        }
        \label{tab:downstream_visa}
    \end{minipage}
\end{table*}

\subsection{Effect of CLIP selection.}
\label{sec:clip_selection}

We evaluate the impact of our CLIP-based quality filter (\cref{sec:clip_filter}) on downstream segmentation performance.
\Cref{tab:clip_ablation} compares models trained on the full set of generated images versus models trained on the CLIP-filtered subset.
Filtering consistently improves image-level AUROC across both benchmarks, while keeping pixel-level AUROC unchanged, confirming that removing misaligned or failed generations yields a cleaner training set for the downstream segmentation model.

\begin{table}[t]
    \centering
    \caption{\textbf{Effect of CLIP-based quality filtering} on downstream anomaly segmentation. Mean image-level AUROC (I) and pixel-level AUROC (P) on MVTec~AD and VisA.}
    \label{tab:clip_ablation}
    \small
    \begin{tabular}{l c >{\columncolor{lightgray!20}}c c >{\columncolor{lightgray!20}}c}
        \toprule
        & \multicolumn{2}{c}{MVTec AD} & \multicolumn{2}{c}{VisA} \\
        \cmidrule(lr){2-3} \cmidrule(lr){4-5}
        Configuration & I & P & I & P \\
        \midrule
        Without CLIP filter & 0.70 & 0.92 & 0.71 & 0.91 \\
        With CLIP filter    & 0.80 & 0.92 & 0.74 & 0.92 \\
        \bottomrule
    \end{tabular}
\end{table}


\section{Conclusion}
\label{sec:conclusion}

We presented \ours{} (Model-agnostic Industrial Realistic Anomaly Generation and Evaluation), a practical, model-agnostic pipeline for industrial anomaly generation that treats the generative model as a black box accessed via API calls, includes a CLIP-based quality filter to retain well-aligned generated images, and combines a lightweight, training-free semantic change detection module for automatic mask creation.

The plug-and-play design means that as stronger generative models or VLMs become available, they can be integrated by simply changing the API endpoint, without re-implementation or retraining. This is a significant advantage over prior methods tightly coupled to specific diffusion architectures. The pipeline requires no task-specific training, no anomalous images, no expensive hardware, and no manual prompt engineering: given a handful of normal reference images, the entire process from defect prompt generation to anomalous image creation is fully automated.

Our extensive benchmark across MVTec~AD and VisA demonstrates that \ours{} produces significantly more realistic anomalous images than all competing generation methods, achieving a TrueSkill score within 0.28 points of real images and the best downstream segmentation performance.
The dual-branch change detection module addresses a critical bottleneck in synthetic anomaly generation by combining text-conditioned Grounding DINO features with fine-grained YOLOv26-Seg structural features, producing high-quality masks without any training and requiring only a small calibration set per category.

Finally, we publicly release a large-scale synthetic anomaly dataset covering all categories of MVTec~AD and VisA (500 image--mask pairs per category, over 13{,}000 pairs in total), together with all generation prompts and code. This is the largest such release to date, and we believe it will be valuable not only for anomaly detection research but also for related tasks such as defect classification and industrial image segmentation.

\paragraph{Limitations and future work.}
Our pipeline depends on the quality and availability of the generative model API; however, the model-agnostic design mitigates this by enabling seamless transition to alternative providers.
The change detection pipeline assumes reasonable visual correspondence between normal and generated images; highly creative generations may reduce mask quality.
Future work will explore extending the pipeline to 3D anomaly detection, few-shot and continual learning settings.

{
    \small
    \bibliographystyle{ieeenat_fullname}
    \bibliography{main}
}

\clearpage
\appendix
\section*{Supplementary Material}

\section{Anomaly Generation Results}
 To demonstrate the generalisation ability and controllability of MIRAGE, Figure~3 presents more generations across different object types and anomaly types; the corresponding descriptions are provided in Table~5.

\begin{table}[htbp]
    \centering
    \caption{\textbf{MVTec AD pixel-level AUC on 1000 ground-truth masks generated by Gemini Image V3.} Results are reported for the \textit{g-dino + yolo} pipeline.}
    \label{tab:mvtec_gemini_masks_auc}
    \large
    \begin{tabular}{lc}
        \toprule
        Category & Pixel-level AUC \\
        \midrule
        bottle & 0.9650 \\
        cable & 0.8866 \\
        capsule & 0.9413 \\
        carpet & 0.9263 \\
        grid & 0.8487 \\
        hazelnut & 0.9718 \\
        leather & 0.9155 \\
        metal\_nut & 0.9405 \\
        pill & 0.9596 \\
        screw & 0.9709 \\
        tile & 0.8955 \\
        toothbrush & 0.9633 \\
        transistor & 0.9643 \\
        wood & 0.9321 \\
        zipper & 0.8594 \\
        \midrule
        \textbf{OVERALL} & \textbf{0.9292} \\
        \bottomrule
    \end{tabular}
\end{table}

\begin{table}[htbp]
    \centering
    \caption{\textbf{VisA pixel-level AUC on 720 ground-truth masks generated by Gemini Image V3.} Results are reported for the \textit{g-dino + yolo} pipeline.}
    \label{tab:visa_gemini_masks_auc}
    \large
    \begin{tabular}{lc}
        \toprule
        Category & Pixel-level AUC \\
        \midrule
        candle & 0.8967 \\
        capsules & 0.8836 \\
        cashew & 0.9459 \\
        chewinggum & 0.8483 \\
        fryum & 0.9553 \\
        macaroni1 & 0.9326 \\
        macaroni2 & 0.9206 \\
        pcb1 & 0.9249 \\
        pcb2 & 0.9448 \\
        pcb3 & 0.9480 \\
        pcb4 & 0.9525 \\
        pipe\_fryum & 0.9700 \\
        \midrule
        \textbf{OVERALL} & \textbf{0.9265} \\
        \bottomrule
    \end{tabular}
\end{table}

\section{Automatic Defect Prompt Generation}
\label{sec:suppl_prompt_gen}

A core component of \ours{} is the fully automatic generation of defect text prompts, which removes the need for manual prompt engineering or prior knowledge of the defect taxonomy for each product category.
For each object category in the dataset, we provide a Vision-Language Model (VLM) with two randomly sampled \emph{normal} (defect-free) reference images and ask it to list 10 realistic defects that could plausibly occur on that object category, each accompanied by a short description suitable as a text prompt for image generation.
By conditioning on actual normal images rather than relying solely on the category name, the VLM leverages its visual understanding of the object's material, geometry, and surface properties to propose contextually grounded defect types (e.g., \emph{scratch}, \emph{dent}, \emph{discoloration}, \emph{crack}).
This procedure is entirely automated, requires no anomalous examples or domain expertise, and is model-agnostic: as more capable VLMs become available, they can be substituted with no change to the pipeline.

The prompt used to generate the defect descriptions is as follows:

\begin{tcolorbox}[colback=white!5!white, colframe=gray!75!black, title=Prompt for Defect Generation]
\small
You are an expert in manufacturing quality control. Given the following two images of a normal, defect-free \{object class\}, please list 10 realistic defects that could plausibly occur on this type of object. 
For each defect, provide a short defect name followed by a description that could be used as a text prompt for image generation. The descriptions should be concise yet specific enough to guide the generation of realistic defect images.
\end{tcolorbox}

The short name of the defect (called ``Defect Type'' in Tab.~\ref{tab:anomaly_descriptions}) is fundamental to condition Grounding DINO for the semantic change detection pipeline, and CLIP for the image selection procedure.


\begin{figure*}[t]
    \centering
    \includegraphics[width=\textwidth]{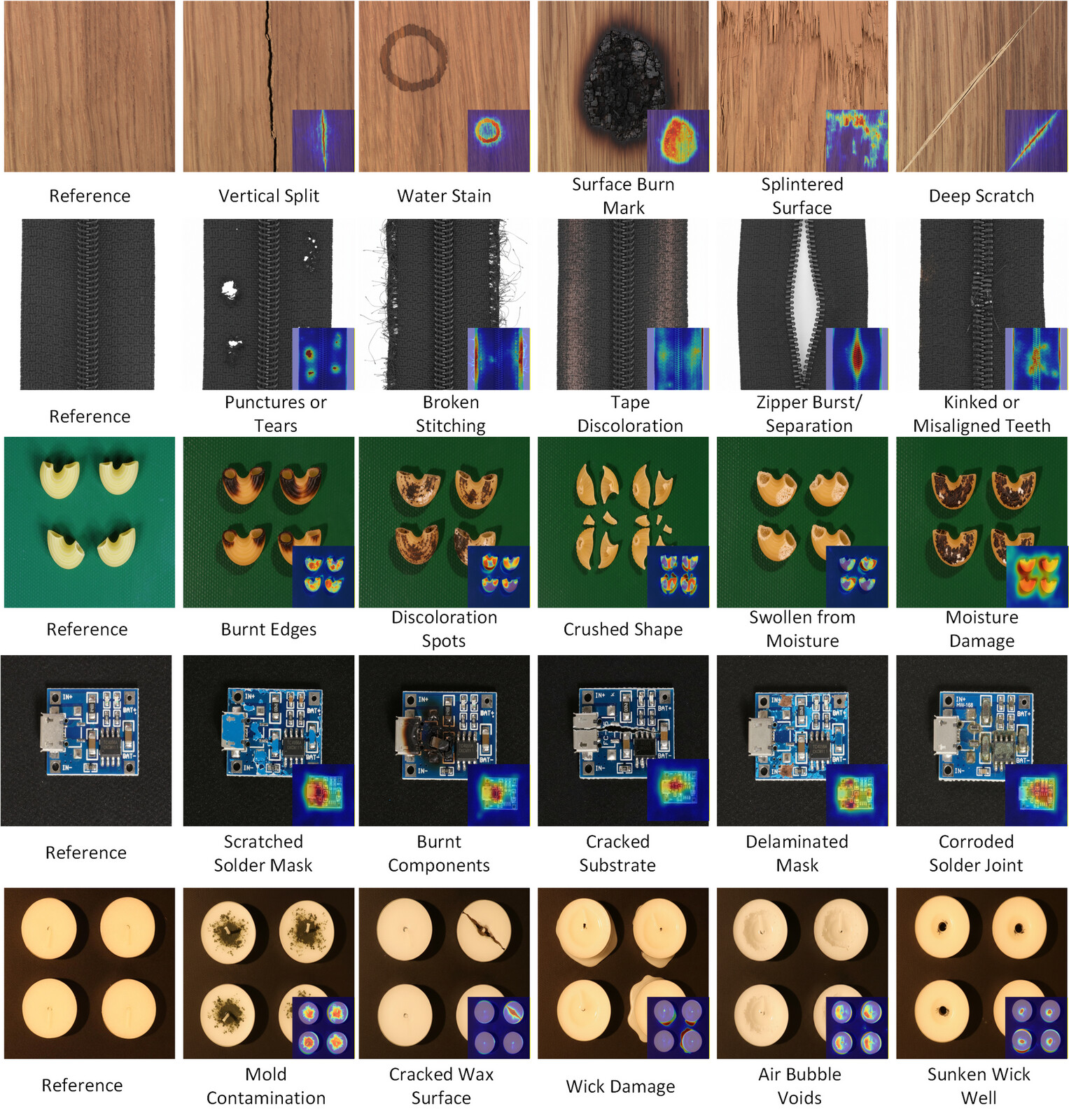}
    \caption{Different types of synthetic anomalous images and corresponding heatmaps generated by MIRAGE.}
    \label{fig:examples_many}
\end{figure*}

\renewcommand{\arraystretch}{1.15}
\begin{longtable}{p{0.12\linewidth} p{0.17\linewidth} p{0.67\linewidth}}
    \caption{\textbf{Corresponding descriptions for anomaly generation} of Fig.~\ref{fig:examples_many}. $\star$ indicates MVTec~AD dataset, while $\star \star$ indicates VisA classes.}
    \label{tab:anomaly_descriptions} \\
    \toprule
    {[Class]} & {[Defect Type]} & GPT-Generated Detailed Descriptions \\
    \midrule
    \endfirsthead
    \multicolumn{3}{l}{\small\itshape (continued from previous page)} \\
    \toprule
    {[Class]} & {[Defect Type]} & GPT-Generated Detailed Descriptions \\
    \midrule
    \endhead
    \midrule
    \multicolumn{3}{r}{\small\itshape (continued on next page)} \\
    \endfoot
    \bottomrule
    \endlastfoot
    \multirow{5}{*}{Wood $\star$}
    & Vertical Split & A deep crack runs parallel to the wood grain, showing a physical separation where the material has dried out and pulled apart, creating a dark void. \\
    & Water Stain & An irregular, dark ring or splotch discolors the wood surface, creating a distinct boundary where moisture has penetrated and altered the grain's appearance.  \\
    & Surface Burn Mark & A dark, charred patch with blackened, carbonized wood fibers disrupts the natural color, indicating contact with a high-heat source.\\
    & Splintered Surface & The surface fibers are raised and peeling away in jagged, needle-like strips, destroying the smooth finish and creating a rough, hazardous texture. \\
    & Deep Scratch & A distinct, light-colored linear gouge cuts across the vertical grain, interrupting the smooth brown pattern where the surface fibers have been severed. \\
    \midrule
    \multirow{5}{*}{Zipper $\star$} & Punctures or Tears & Visible holes or rips in the woven fabric tape, distinct from the edges. \\
    & Broken Stitching & The specific thread running along the edge of the coil snaps or unravels, causing the plastic coil to detach from the fabric tape. \\
    & Tape Discoloration & The black color of the fabric fades to gray or reddish-brown due to UV (sun) exposure or chemical bleaching. \\
    & Zipper Burst/Separation & The left and right sides of the coil separate and pop open in the middle, even though the slider (not pictured) might be pulled up. \\
    & Kinked or Misaligned Teeth & The smooth, straight line of the center zipper chain looks crooked, bent, or has a distinct 'hump,' preventing the slider from moving past that point. \\
    \midrule
    \multirow{5}{*}{Macaroni 1 $\star \star$} & Burnt Edges & The rims and tips of the shell appear darkly toasted or charred, turning brown or black from excessive heat during drying or storage. \\
    & Discoloration Spots & Irregular dark brown, black, or white patches mar the uniform yellow-orange color of the pasta, indicating mold, burning, or contamination. \\
    & Crushed Shape & The tubular form is flattened or compressed, losing the round hollow cross-section and characteristic elbow curvature. \\
    & Swollen from Moisture & The pasta appears bloated and soft rather than hard and dry, with the surface losing its smooth finish and becoming sticky or tacky. \\
    & Moisture Damage & The pasta appears swollen, soft, or sticky rather than dry and hard, with a distorted shape and loss of the crisp texture from humidity exposure. \\
    \midrule
    \multirow{5}{*}{PCB 4 $\star \star$} & Scratched Solder Mask & Deep gouges or scrapes cut through the blue protective coating, exposing the underlying copper or substrate material in irregular patterns. \\
    & Burnt Components & The ultrasonic sensor mesh, resistors, or solder joints show blackened, charred areas with blistered plastic indicating electrical overcurrent or short circuit damage. \\
    & Cracked Substrate & A visible fracture line runs through the blue PCB material, splitting the board and potentially severing internal copper traces beneath the surface. \\
    & Delaminated Mask & The blue solder mask protective coating is peeling, bubbling, or separated from the PCB surface in patches. \\
    & Corroded Solder Joint & The solder connection appears dull, rough, or covered with dark oxidation, indicating poor contact and potential circuit failure. \\
    \midrule
    \multirow{5}{*}{Candle $\star \star$} & Mold Contamination & Small, dark green or black fuzzy spots appear scattered across the wax surface, indicating fungal growth from moisture exposure in storage. \\
    & Cracked Wax Surface & The smooth cream-colored wax top features a jagged fissure running from the edge toward the center wick hole, exposing a darker interior layer where the material has split under thermal stress. \\
    & Wick Damage & The central fabric wick is frayed, broken off at the base, or completely missing from its anchoring hole, leaving an empty void or torn fiber ends. \\
    & Air Bubble Voids & Large spherical cavities appear trapped beneath the smooth wax surface, visible as bulging domes or collapsed pits that break the uniform flat plane of the candle top. \\
    & Sunken Wick Well & The circular depression surrounding the wick appears excessively deep and irregular, with the wax collapsing inward to create an uneven crater that disrupts the flat top surface. \\
\end{longtable}

\end{document}